\documentclass[conference,11pt]{IEEEtran}
\IEEEoverridecommandlockouts
\usepackage{cite}
\usepackage{amsmath,amssymb,amsfonts}
\usepackage{algorithmic}
\usepackage{graphicx}
\usepackage{textcomp}
\usepackage{xcolor}
\def\BibTeX{{\rm B\kern-.05em{\sc i\kern-.025em b}\kern-.08em
    T\kern-.1667em\lower.7ex\hbox{E}\kern-.125emX}}
\usepackage[utf8]{inputenc}
\usepackage[T1]{fontenc}
\usepackage{lmodern}
\usepackage{balance}
\usepackage{subcaption}
\usepackage{booktabs}
\usepackage{multirow}
\usepackage[hidelinks]{hyperref}
\usepackage{tikz}
\usetikzlibrary{arrows.meta, positioning, shapes.geometric, calc}

\begin{document}

\title{\fontsize{20pt}{\baselineskip}\selectfont Is Domain Adaptation Always Helpful?\\A Frozen-Backbone Study of Cross-Domain Sentiment Transfer}

\author{
Phat Tran \qquad
Artin Lahni \qquad
Pranav Kulkarni \qquad
Yaolun Zhang \\
Oregon State University\\
{\tt\small \{tranphat, lahnia, kulkarnp, zhanyaol\}@oregonstate.edu}
}
\maketitle
\thispagestyle{plain}
\pagestyle{plain}

\begin{abstract}
    Sentiment analysis with frozen pre-trained language model (PLM) backbones has become a common paradigm, yet the practical benefit of explicit domain adaptation remains unclear, particularly when backbones encode varying degrees of target-domain knowledge. We present a preliminary case study evaluating a controlled family of frozen embedding backbones (Qwen3-Embedding 0.6B, 4B, 8B), alongside RoBERTa-base and FinBERT. We train a lightweight MLP adapter on consumer reviews using Domain-Adversarial Neural Networks (DANN), Maximum Mean Discrepancy (MMD), and Supervised Contrastive Learning (SCL), and evaluate transfer to movie reviews (SST-2) and a heavily restricted subset of financial news (Financial PhraseBank). Within this constrained sample, we observe two distinct transfer patterns. On SST-2, domain adaptation provides negligible gain regardless of scale. On the financial subset, explicit domain adaptation appears to recover substantial performance for small general-purpose backbones. Notably, we find that adversarial alignment (DANN) is associated with degraded performance for domain-specialized backbones like FinBERT, consistent with erosion of pre-existing domain-specific structure, whereas supervised contrastive loss appears to preserve it. These preliminary findings suggest that the efficacy of explicit domain adaptation is highly contingent on whether the frozen backbone already possesses target-domain coverage.
\end{abstract}


\section{Introduction}
\label{sec:intro}
Sentiment analysis, the task of classifying the subjective polarity of text, is a cornerstone application of natural language processing (NLP), powering everything from product recommendation systems to financial market monitoring. Modern approaches increasingly rely on large pre-trained language models (PLMs) used as frozen feature extractors, with lightweight task heads trained on top of their embeddings~\cite{araci2019finbert,liu2019roberta}. As these backbone models have scaled from hundreds of millions to billions of parameters, their reported performance on standard benchmarks has grown correspondingly impressive \cite{choi2025how}.

However, this progress raises a critical measurement problem: at scale, PLMs may be exposed to benchmark evaluation examples during pretraining, making it difficult to distinguish genuine generalization from memorization. When a frozen backbone has already seen SST-2~\cite{socher2013sst} or Financial PhraseBank~\cite{malo2014financial} during pretraining, high zero-shot accuracy reflects data contamination rather than genuine generalization. This makes it fundamentally unclear whether explicit domain adaptation techniques, methods that align a model's representations across source and target domains, are truly helpful, or whether they merely compensate for contamination artifacts introduced by smaller, less-contaminated backbones.

We address this question directly: ``When does domain adaptation help for sentiment transfer?'' To reduce reliance on potentially contaminated benchmark memorization, we use the Qwen3-Embedding series (0.6B, 4B, and 8B parameters)~\cite{qwen3embedding}, an embedding family that provides a controlled scale progression for studying frozen-backbone transfer. This yields a setting in which the effects of backbone scale and domain specialization can be compared more directly. On top of these frozen embeddings, we train a lightweight MLP adapter using three complementary domain adaptation objectives: Domain-Adversarial Neural Networks (DANN)~\cite{ganin2016dann}, Maximum Mean Discrepancy (MMD)~\cite{gretton2012kernel}, and Supervised Contrastive Learning (SCL)~\cite{khosla2020supcon}. We train on consumer product and restaurant reviews (Yelp Reviews~\cite{zhang2015yelp} and Amazon Polarity~\cite{mcauley2013amazon,zhang2015yelp}) and evaluate zero-shot transfer to movie reviews (SST-2~\cite{socher2013sst,wang2018glue}) and financial news (Financial PhraseBank~\cite{malo2014financial}), with no target labels used during training.

Our experiments reveal two qualitatively distinct transfer regimes, clarifying when domain adaptation is and is not worth applying. On SST-2, which shares informal opinionated language with the source domains, all backbone configurations achieve strong zero-shot performance (macro F$_1$ of 0.85 to 0.91), and adding any combination of DA losses provides negligible additional benefit. On Financial PhraseBank, however, the domain shift is severe: the 0.6B backbone without DA achieves only 0.309 F$_1$, while the full DANN and MMD combination recovers 0.637. Critically, we find that a domain-specialized backbone (FinBERT) reaches 0.902 F$_1$ without any adaptation, but its performance is actively harmed by adversarial alignment (DANN), which erodes the domain-specific structure that makes it valuable. Together, these findings suggest a unified principle: explicit domain adaptation is necessary and effective only when the frozen backbone lacks coverage of the target distribution, and it becomes harmful when applied against a backbone whose pretraining already encodes target domain structure.

This work makes the following contributions:
\begin{itemize}
    \item We present an empirical case study of domain adaptation with frozen sentiment backbones using Qwen3-Embedding models of different scales, together with RoBERTa and FinBERT baselines.
    \item We observe that adaptation has little measurable effect on close-domain transfer to SST-2, but substantially improves cross-domain transfer to Financial PhraseBank in our setting, with gains up to 32.8 macro F$_1$ points on a 93-example evaluation set.
    \item We find that adversarial alignment can harm domain-specialized backbones such as FinBERT, while supervised contrastive learning achieves the best financial result observed in our experiments.
    \item We propose a preliminary, exploratory guideline: use distribution-matching adaptation when the backbone lacks target-domain coverage, and use contrastive refinement when target-domain structure is already present, pending validation on larger benchmarks.
\end{itemize}
\section{Related Work}
\label{sec:related}
\subsection{Sentiment Analysis with Pre-trained Language Models}
Early neural approaches to sentiment analysis relied on task-specific architectures trained from scratch, but the field shifted decisively toward transferring representations from large pre-trained language models~\cite{liu2019roberta}. A common instantiation uses BERT-style encoders as frozen feature extractors, where a simple probe or classification head is trained on top of fixed representations~\cite{tenney2019bert}. While this probing paradigm reveals what information is encoded in pretrained representations, it also introduces a measurement risk: as model scale grows, benchmark examples may appear in pretraining corpora, making it difficult to distinguish genuine generalization from memorization~\cite{deng2023benchmark, xu2024benchmarkdatacontaminationlarge, sun2025the}. Our work is motivated by this risk. We use a controlled family of frozen Qwen3-Embedding backbones at multiple scales, together with RoBERTa and FinBERT baselines, to study when domain adaptation improves transfer under fixed representations.

\subsection{Unsupervised Domain Adaptation for NLP}
Unsupervised domain adaptation (UDA) for NLP addresses the challenge of transferring a model trained on labeled source text to an unlabeled target domain~\cite{ramponi2020survey}. The dominant family of approaches encourages domain-invariant representations through adversarial training, of which DANN~\cite{ganin2016dann} is the canonical example, using a gradient reversal layer to make the feature extractor indistinguishable across domains. A complementary family uses distribution-matching losses such as MMD~\cite{gretton2012kernel}, which minimizes a kernel-based distance between source and target feature distributions without requiring an explicit discriminator. However, theoretical analysis of MMD for domain adaptation has shown that naive distribution matching can degrade intra-class discriminability~\cite{wang2020rethink}, a finding consistent with our own ablation results showing that adding contrastive loss on top of DANN and MMD hurts performance for small-capacity backbones. Crucially, all of these prior works apply adaptation while jointly fine-tuning the backbone, making it difficult to disentangle adaptation signal from backbone drift \cite{qin2024freeze}. Our two-stage frozen pipeline isolates this variable by design.

\subsection{Contrastive Learning for Representation Alignment}
SCL~\cite{khosla2020supcon} trains representations by pulling same-class samples together and pushing apart cross-class pairs using temperature-scaled cosine similarity. When applied to domain adaptation, contrastive objectives are appealing because they can sharpen class boundaries without requiring target labels \cite{simoncini2024no}. Prior work on aspect-based sentiment adaptation~\cite{wiseda2022aspect} shows that combining adversarial and contrastive losses improves robustness to domain shift. We build on this intuition but uncover an important asymmetry: for domain-specialized backbones such as FinBERT, contrastive loss is the single most effective component (+0.076 macro F$_1$), while DANN appears to suppress domain-specific information that makes those representations valuable. This interaction between backbone specialization and adaptation-objective type has not been characterized in prior frozen-backbone settings.

\subsection{Pseudo-Labeling in Semi-Supervised Adaptation}
Pseudo-labeling augments supervised training with high-confidence model predictions on unlabeled target data, providing a weak supervision signal across domain boundaries. Chen et al.~\cite{chen2021pseudolabel} show that pseudo-label guided UDA of contextual embeddings outperforms strong baselines on both named entity recognition and sentiment analysis, using masked language modeling on target text to bootstrap label assignment. Compared to that approach, we treat pseudo-labeling as a supplementary signal activated at epochs 4 and 5 (of 5 total) with a high confidence threshold (0.95), layered on top of the primary DANN, MMD, and contrastive objectives rather than as a standalone adaptation mechanism. This conservative design ensures the adapter has converged sufficiently on source data before target predictions become reliable enough to incorporate, preventing confirmation bias from corrupting adaptation in early training.

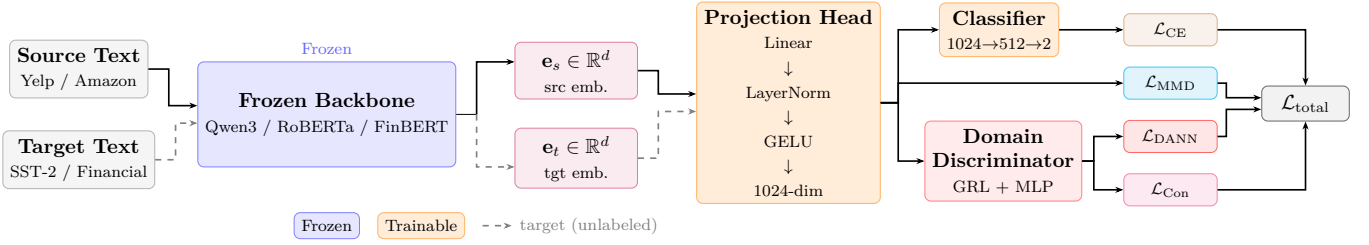
\begin{figure*}[!htbp]
    \centering
    \resizebox{\textwidth}{!}{
        \begin{tikzpicture}[
    node distance=0.55cm and 0.7cm,
    box/.style={rectangle, rounded corners=3pt, minimum width=2.1cm, minimum height=0.65cm, align=center, font=\small},
    frozen/.style={box, fill=blue!10, draw=blue!50},
    trainable/.style={box, fill=orange!15, draw=orange!60},
    loss/.style={box, minimum width=1.6cm, minimum height=0.55cm, font=\scriptsize},
    arrow/.style={-{Stealth[length=4pt]}, thick},
    darrow/.style={-{Stealth[length=4pt]}, thick, dashed, gray},
]

\node[box, draw=gray!60, fill=gray!8] (src) {\textbf{Source Text}\\{\scriptsize Yelp / Amazon}};
\node[box, draw=gray!60, fill=gray!8, below=of src] (tgt) {\textbf{Target Text}\\{\scriptsize SST-2 / Financial}};

\node[frozen, minimum height=1.8cm, minimum width=2.3cm] (backbone) at ($(src)!0.5!(tgt)+(4.25cm,0)$) {\textbf{Frozen Backbone}\\{\scriptsize Qwen3 / RoBERTa / FinBERT}};

\node[box, draw=purple!50, fill=purple!8, right=1cm of backbone, yshift=0.75cm] (semb) {$\mathbf{e}_s \in \mathbb{R}^d$\\{\scriptsize src emb.}};

\node[box, draw=purple!50, fill=purple!8, right=1cm of backbone, yshift=-0.75cm] (temb) {$\mathbf{e}_t \in \mathbb{R}^d$\\{\scriptsize tgt emb.}};

\node[trainable, right=1cm of semb, yshift=-0.55cm, minimum height=1.8cm, minimum width=2.4cm] (proj) {\textbf{Projection Head}\\{\scriptsize Linear}\\{\scriptsize $\downarrow$}\\{\scriptsize LayerNorm}\\{\scriptsize $\downarrow$}\\{\scriptsize GELU}\\{\scriptsize $\downarrow$}\\{\scriptsize 1024-dim}};

\node[trainable, right=1.0cm of proj, yshift=1.25cm, minimum width=2.0cm] (cls) {\textbf{Classifier}\\{\scriptsize 1024$\to$512$\to$2}};

\node[trainable, right=0.75cm of proj, yshift=-1.0cm, minimum width=2.0cm, draw=red!50, fill=red!8] (disc) {\textbf{Domain}\\\textbf{Discriminator}\\{\scriptsize GRL + MLP}};

\node[loss, right=1.1cm of cls, fill=brown!10, draw=brown!60] (lce) {$\mathcal{L}_{\text{CE}}$};

\node[loss, below=0.35cm of lce, fill=cyan!10, draw=cyan!60] (lmmd) {$\mathcal{L}_{\text{MMD}}$};

\node[loss, below=0.35cm of lmmd, fill=red!10, draw=red!60] (ldann) {$\mathcal{L}_{\text{DANN}}$};

\node[loss, below=0.35cm of ldann, fill=magenta!10, draw=magenta!60] (lcon) {$\mathcal{L}_{\text{Con}}$};

\node[box, draw=black!60, fill=black!5, minimum width=1.5cm, minimum height=0.6cm, right=0.75cm of lmmd, yshift=-0.35cm, font=\small] (ltot) {$\mathcal{L}_{\text{total}}$};

\node[font=\scriptsize, text=blue!60, above=0cm of backbone] {Frozen};

\draw[arrow] (src.east) -- ++(0.35,0) |- ([yshift=0.15cm]backbone.west);
\draw[darrow] (tgt.east) -- ++(0.35,0) |- ([yshift=-0.15cm]backbone.west);

\coordinate (bb_out) at (backbone.east);
\draw (bb_out) -- ++(0.35,0) coordinate (split);

\draw[arrow] (split) |- ([yshift=0.15cm]semb.west);
\draw[darrow] (split) |- ([yshift=-0.15cm]temb.west);

\draw[arrow] (semb.east) -- ++(0.35,0) |- ([yshift=0.15cm]proj.west);
\draw[darrow] (temb.east) -- ++(0.35,0) |- ([yshift=-0.15cm]proj.west);

\draw[arrow] (proj.east) -- ++(0.3,0) |- (cls.west);
\draw[arrow] (proj.east) -- ++(0.3,0) |- (disc.west);
\draw[arrow] (proj.east) -- ++(0.3,0) |- (lmmd.west);

\draw[arrow] (cls.east) -- ++(0.2,0) |- (lce.west);
\draw[arrow] (disc.east) -- ++(0.2,0) |- (ldann.west);
\draw[arrow] (disc.east) -- ++(0.2,0) |- (lcon.west);

\draw[arrow] (lce.east) -- ++(0.25,0) -| (ltot.north);
\draw[arrow] (lmmd.east) -- ++(0.15,0) |- ([yshift=0.1cm]ltot.west);
\draw[arrow] (ldann.east) -- ++(0.15,0) |- ([yshift=-0.1cm]ltot.west);
\draw[arrow] (lcon.east) -- ++(0.25,0) -| (ltot.south);

\node[frozen, minimum width=1.1cm, minimum height=0.4cm, font=\scriptsize] at ([yshift=-1cm]backbone.south) (leg1) {Frozen};

\node[trainable, minimum width=1.1cm, minimum height=0.4cm, font=\scriptsize, right=0.3cm of leg1] (leg2) {Trainable};

\draw[darrow] ([xshift=0.3cm]leg2.east) -- ++(0.5,0) node[right, font=\scriptsize, text=gray] {target (unlabeled)};

\end{tikzpicture}
    }
    \caption{Overview of the frozen-backbone domain adaptation pipeline. Source and target texts are encoded by a fixed pretrained backbone and mapped into a shared adapter space. The classifier is trained with source-label cross-entropy, while DANN, MMD, and supervised contrastive losses encourage domain alignment and class separation. Only the adapter, classifier, and domain discriminator are updated during training.}
    \label{fig:pipeline}
\end{figure*}

\section{Methodology}

We adopt a two-stage pipeline that decouples representation learning from domain adaptation. A frozen pre-trained backbone encodes each input text into a fixed-length vector. A lightweight adapter is then trained on top of these frozen embeddings using labeled source data and domain adaptation losses applied to unlabeled target samples. Fig.~\ref{fig:pipeline} illustrates the overall pipeline. Target-domain embeddings are used with stop-gradient to ensure that domain-alignment losses do not update the frozen backbone.

\subsection{Frozen Backbone}
Freezing the backbone is a deliberate design choice: it eliminates representation drift during adaptation and allows performance differences across domains to be attributed to the adapter and domain-adaptation objectives rather than to changes in the backbone itself. We evaluate five backbone configurations. From the Qwen embedding family, we include Qwen3-Embedding-8B (8B parameters, 4096-dimensional output), Qwen3-Embedding-4B (4B parameters, 2560-dimensional output), and Qwen3-Embedding-0.6B (0.6B parameters, 1024-dimensional output) \cite{qwen3embedding}. In addition, we consider two widely used encoder baselines: RoBERTa-base (125M parameters, 768-dimensional output)~\cite{liu2019roberta} and FinBERT (110M parameters, 768-dimensional output)~\cite{araci2019finbert}, a BERT variant further pre-trained on financial corpora. All backbones are kept fully frozen during training. Embeddings are pre-computed once per dataset split and cached, so no gradients flow into the backbone at any stage.

\subsection{Adapter Architecture}
The adapter is a two-layer MLP projection head followed by a linear classifier, with approximately 6.3M trainable parameters, regardless of the backbone. The projection maps the backbone embedding dimension to a shared 1024-dimensional space via two blocks of \texttt{Linear} $\to$ \texttt{LayerNorm} $\to$ \texttt{GELU} $\to$ \texttt{Dropout(0.1)}. The classifier maps to 512 dimensions before producing binary logits. A separate domain discriminator, consisting of \texttt{Linear(1024\,$\to$\,512)} $\to$ \texttt{ReLU} $\to$ \texttt{Dropout(0.1)} $\to$ \texttt{Linear(512\,$\to$\,K)}, is used for the DANN objective.

\subsection{Domain Adaptation Losses}
We study three domain adaptation objectives combined with standard cross-entropy. The base supervised objective is standard cross-entropy on labeled source samples:
\begin{equation}
    \label{eq:ce}
    \mathcal{L}_{\text{CE}} =
    -\frac{1}{n}\sum_{i=1}^{n}
    \sum_{c=1}^{C} y_{ic}\,\log\hat{y}_{ic}
\end{equation}
where $y_{ic}$ is the ground-truth label indicator and $\hat{y}_{ic}$ is the
predicted probability for class $c \in \{0,1\}$.

\paragraph{DANN}
A gradient reversal layer (GRL) negates gradients flowing from the domain discriminator into the feature extractor, encouraging domain-invariant representations:
\begin{equation}
    \mathcal{L}_{\text{DANN}} =
    \frac{1}{n}\sum_{i=1}^{n}
    \mathcal{L}_{\text{dom}}\!\left(D\!\left(\text{GRL}(f_i)\right),\, d_i\right)
\end{equation}
where $f_i$ is the projected feature for sample $i$, $d_i \in \{0,\dots,K{-}1\}$ is its domain label, $D(\cdot)$ is the domain discriminator, and $\text{GRL}$ negates gradients during backpropagation with scale $\lambda_{\text{DANN}}$, which warms up linearly over the first 500 steps to prevent early instability. The inner loss follows the same cross-entropy form as Equation (\ref{eq:ce}), applied over $K$ domain classes rather than sentiment classes.

\paragraph{MMD}
Maximum mean discrepancy with a Gaussian kernel ($\sigma{=}1.0$) minimizes the distance between source and target feature distributions at each training step:
\begin{equation}
    \mathcal{L}_{\text{MMD}}(S, T) =
    \left\|
    \frac{1}{n}\sum_{i=1}^{n}\phi(s_i)
    -
    \frac{1}{m}\sum_{j=1}^{m}\phi(t_j)
    \right\|^2_{\mathcal{H}}
\end{equation}
where $\phi$ is the feature map induced by the Gaussian kernel, $S$ and $T$ are the source and target feature sets, and
$n$, $m$ are their respective batch sizes.

\paragraph{Supervised Contrastive Loss}
Same-class feature pairs are pulled together and cross-class pairs pushed apart using temperature-scaled cosine similarity ($\tau{=}0.07$), using source labels only:
\begin{equation}
    \mathcal{L}_{\text{Con}} =
    \sum_{i \in I}
    \frac{-1}{|P(i)|}
    \sum_{p \in P(i)}
    \log
    \frac{
        \exp\!\left(\mathbf{z}_i \cdot \mathbf{z}_p \,/\, \tau\right)
    }{
        \sum_{a \neq i} \exp\!\left(\mathbf{z}_i \cdot \mathbf{z}_a \,/\, \tau\right)
    }
\end{equation}
where $P(i)$ is the set of same-class indices in the batch, $\mathbf{z}$ are $\ell_2$-normalized features from the projection head, and $\tau = 0.07$ is the temperature.

The total loss is:
\begin{equation}
    \mathcal{L} = \mathcal{L}_{\text{CE}} + \lambda_{\text{DANN}}\,\mathcal{L}_{\text{DANN}} + \lambda_{\text{MMD}}\,\mathcal{L}_{\text{MMD}} + \lambda_{\text{Con}}\,\mathcal{L}_{\text{Con}}
\end{equation}

\subsection{Pseudo-Labeling}
At epochs 4 and 5 (of 5 total), high-confidence target predictions (softmax confidence $\geq 0.95$) are incorporated into training with their predicted labels, providing weak supervision on the target domain. This conservative schedule ensures the adapter has converged sufficiently on source data before target predictions become reliable enough to incorporate, preventing confirmation bias from corrupting adaptation in early training.

\subsection{Training Details}
All adapters are trained for 5 epochs with AdamW ($\text{lr}{=}2{\times}10^{-4}$, weight decay $0.01$)~\cite{adamw2017}, cosine learning rate schedule with 6\% warmup, gradient clipping at 1.0, and bfloat16 mixed precision. Batch size is 32 with gradient accumulation over 2 steps. Following standard practice in multi-objective domain adaptation~\cite{ganin2016dann,gretton2012kernel}, the loss weights are set to $\lambda_{\text{DANN}}{=}0.1$, $\lambda_{\text{MMD}}{=}0.05$, and $\lambda_{\text{Con}}{=}0.05$, balancing the contribution of each alignment objective relative to the primary cross-entropy supervision signal. All experiments use a fixed random seed of 42 and run on a single NVIDIA GPU.

\begin{figure}[htbp]
    \centering
    \includegraphics[width=\columnwidth]{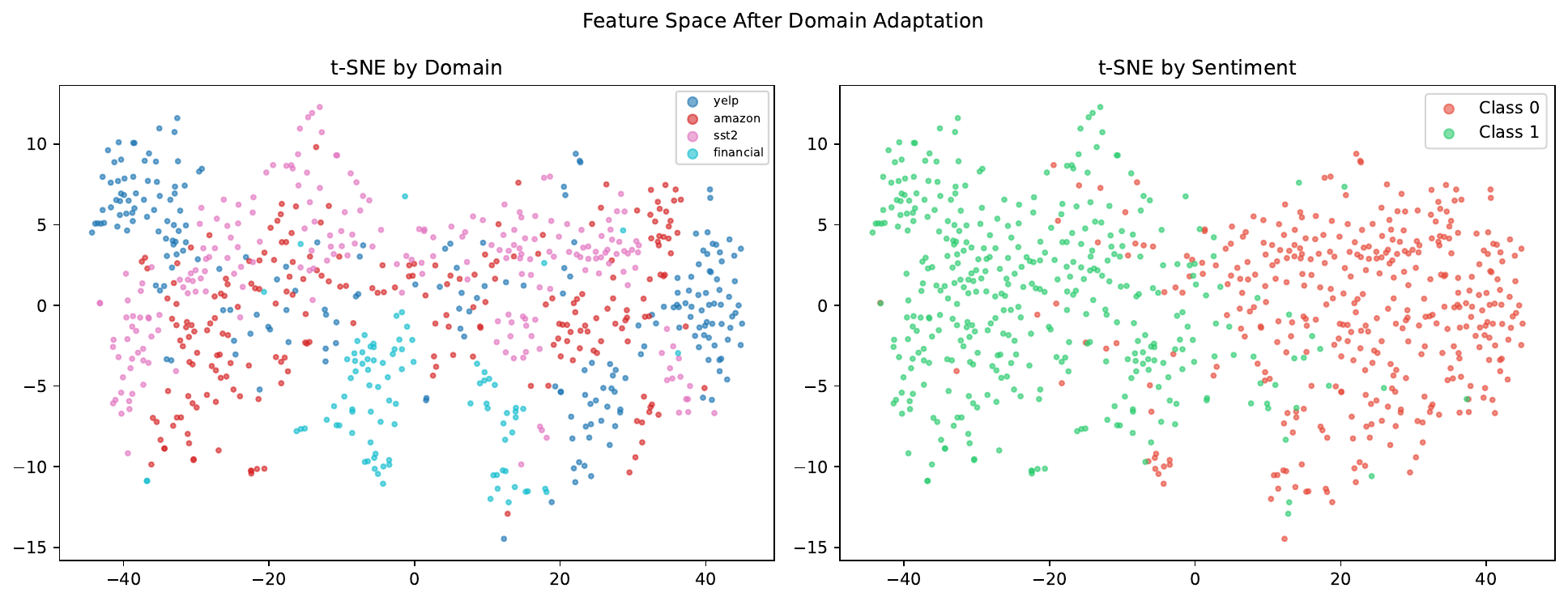}
    \caption{t-SNE of adapter features (Qwen3-0.6B), colored by domain (left) and sentiment (right), showing domain-mixed yet class-separated representations.}
    \label{fig:tsne}
\end{figure}

\section{Experiments}

\subsection{Setup}

\paragraph{Datasets}
We train on two labeled source domains (Yelp Reviews and Amazon Polarity) and evaluate zero-shot transfer to two target domains (SST-2 and Financial PhraseBank), with no target labels used during training. Dataset statistics are shown in Table~\ref{tab:datasets}. We convert Yelp and Financial PhraseBank to binary sentiment by excluding their neutral classes: 3-star reviews for Yelp and neutral examples for Financial PhraseBank.

We note that DANN, MMD, and pseudo-labeling draw unlabeled target features from the same pool of examples later used for zero-shot evaluation, since Table~\ref{tab:datasets} shows no separate target training split. This is standard practice in transductive unsupervised domain adaptation, but it means our reported ``zero-shot'' results reflect transductive rather than inductive generalization, and should not be interpreted as performance on data unseen at training time.

\begin{table*}
    \centering
    \normalsize
    \begin{tabular}{@{}llrrcrl@{}}
        \toprule
        \multicolumn{1}{c}{\textbf{Domain}}    &
        \multicolumn{1}{c}{\textbf{Role}}      &
        \multicolumn{1}{c}{\textbf{Train}}     &
        \multicolumn{1}{c}{\textbf{Eval}}      &
        \multicolumn{1}{c}{\textbf{Pos/Neg (\%)}}   &
        \multicolumn{1}{c}{\textbf{Avg.\ Len}} &
        \multicolumn{1}{c}{\textbf{Source}}                                                                                        \\
        \midrule
        Yelp Reviews                           & Source & 31,880 & 3,997 & 46 / 54 & 128 & \cite{zhang2015yelp}                    \\
        Amazon Polarity                        & Source & 40,000 & 5,000 & 51 / 49 & 76  & \cite{mcauley2013amazon, zhang2015yelp} \\
        SST-2                                  & Target & ---    & 872   & 51 / 49 & 20  & \cite{socher2013sst}                    \\
        Financial PhraseBank                   & Target & ---    & 93    & 61 / 39 & 24  & \cite{malo2014financial}                \\
        \bottomrule
    \end{tabular}
    \caption{Dataset statistics. Yelp counts are after binarization, with 3-star neutral reviews removed. Financial PhraseBank counts are after removing the neutral class. Pos/Neg shows class balance (\%). Avg.\ Len is the mean word count per sample. Target labels are held out during training.}
    \label{tab:datasets}
\end{table*}
\paragraph{Main Pipeline vs. Ablation Protocol}
For the main results in Table~\ref{tab:main}, we train the full pipeline with all three DA losses. We first use SST-2 as the adaptation target and then adapt to Financial PhraseBank, following a curriculum intended to stabilize training on the harder financial domain.

For the ablation study (\S\ref{sec:ablation}), each DA configuration is trained from scratch with Financial PhraseBank as the sole target domain. This provides a cleaner measure of each component's individual contribution.
\subsection{Main Results}

Table~\ref{tab:main} reports macro F$_1$ for all backbone--domain combinations under the full DA pipeline.

\begin{table*}[t]
    \centering
    \normalsize
    \begin{tabular}{@{}lc rrrr rrrr@{}}
        \toprule
        \multicolumn{1}{@{}c}{\multirow{2}{*}{\textbf{Model}}} & \multirow{2}{*}{\textbf{Params}} & \multicolumn{4}{c}{\textbf{Macro F$_1$}} & \multicolumn{4}{c}{\textbf{Accuracy}}                                                                                                                                                                                                                                \\
        \cmidrule(lr){3-6} \cmidrule(l){7-10}
                                                               &                                  & \multicolumn{1}{c}{\textbf{Yelp}}        & \multicolumn{1}{c}{\textbf{Amazon}}   & \multicolumn{1}{c}{\textbf{SST-2}} & \multicolumn{1}{c}{\textbf{Fin.}} & \multicolumn{1}{c}{\textbf{Yelp}} & \multicolumn{1}{c}{\textbf{Amazon}} & \multicolumn{1}{c}{\textbf{SST-2}} & \multicolumn{1}{c@{}}{\textbf{Fin.}} \\
        \midrule
        Qwen3-Emb-0.6B                                         & 0.6B                             & 0.953                                    & 0.929                                 & 0.893                              & 0.689                             & 0.953                             & 0.929                               & 0.893                              & 0.699                                \\
        Qwen3-Emb-4B                                           & 4B                               & 0.972                                    & 0.948                                 & 0.908                              & 0.793                             & 0.972                             & 0.948                               & 0.908                              & 0.796                                \\
        Qwen3-Emb-8B                                           & 8B                               & 0.976                                    & 0.950                                 & 0.913                              & 0.944                             & 0.976                             & 0.950                               & 0.913                              & 0.946                                \\
        RoBERTa-base                                           & 125M                             & 0.951                                    & 0.918                                 & 0.848                              & 0.748                             & 0.951                             & 0.918                               & 0.849                              & 0.774                                \\
        FinBERT                                                & 110M                             & 0.921                                    & 0.891                                 & 0.854                              & 0.731                             & 0.922                             & 0.891                               & 0.854                              & 0.731                                \\
        \bottomrule
    \end{tabular}
    \caption{Macro F$_1$ and accuracy across all backbone--domain combinations in the full DA pipeline. Source domains use in-domain test splits, and target domains are evaluated in a zero-shot setting.}
    \label{tab:main}
\end{table*}

Results reveal two distinct transfer regimes. On SST-2, performance is stable across all models (0.85--0.91), with the gap between the smallest and largest models only 6.5 points. Consumer reviews and movie reviews share informal, opinionated language, so even compact backbones generalize effectively. The financial domain shows a different pattern: Qwen3-Embedding models span a 25.5-point range (0.689 to 0.944) that tracks closely with model scale, while RoBERTa (0.748) and FinBERT (0.731) fall in between, despite being substantially smaller than the Qwen3-Embedding models.

Notably, FinBERT underperforms RoBERTa on the financial domain under the full pipeline despite its financial pretraining. As the ablation study suggests, this is not due simply to a lack of financial representations, since its no-DA baseline performs considerably better, but rather because the combined domain adaptation losses in the full pipeline, particularly DANN, degrade the domain-specific features that make FinBERT effective.

\begin{figure}[t]
    \centering
    \includegraphics[width=\columnwidth]{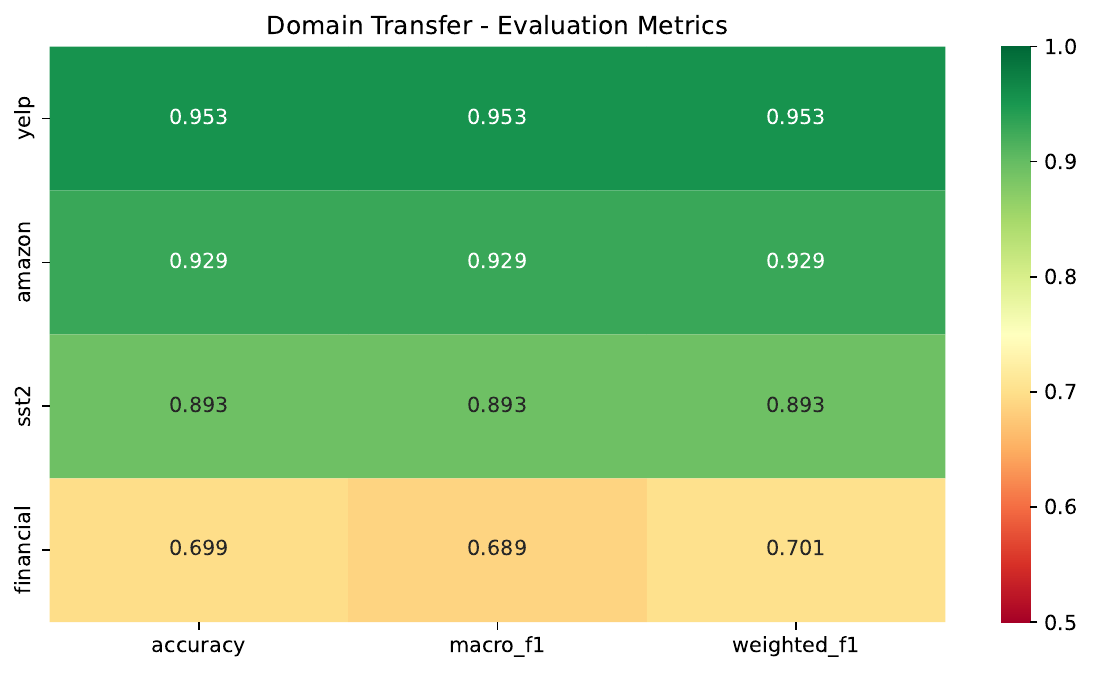}
    \caption{Domain transfer heatmap for Qwen3-0.6B, with rows as source domains and columns as evaluation domains. Off-diagonal entries show zero-shot performance.}
    \label{fig:heatmap}
\end{figure}

\subsection{Ablation Study}
\label{sec:ablation}

To isolate each component's contribution, we train the adapter once for each of six DA configurations per backbone, targeting only the financial domain from scratch. Table~\ref{tab:ablation_financial} reports macro F$_1$ on Financial PhraseBank. Because this evaluation set contains only 93 examples, single-run macro F$_1$ differences of a few points should be interpreted with caution. We report multi-seed variance only for the 8B backbone due to compute constraints, and treat other single-run comparisons in this section as suggestive rather than conclusive.

\begin{table*}[t]
    \centering
    \normalsize
    \begin{tabular}{@{}lccccccc@{}}
        \toprule
        \multicolumn{1}{c}{\textbf{Condition}}    &
        \multicolumn{1}{c}{\textbf{D}}            &
        \multicolumn{1}{c}{\textbf{M}}            &
        \multicolumn{1}{c}{\textbf{C}}            &
        \multicolumn{1}{c}{\textbf{0.6B}}         &
        \multicolumn{1}{c}{\textbf{4B}}           &
        \multicolumn{1}{c}{\textbf{8B}$^\dagger$} &
        \multicolumn{1}{c}{\textbf{FinBERT}}                                                                                                                 \\
        \midrule
        Baseline                                  &            &            &            & 0.309          & 0.632          & 0.421          & 0.902          \\
        + DANN                                    & \checkmark &            &            & 0.389          & 0.708          & \textbf{0.428} & 0.796          \\
        + MMD                                     &            & \checkmark &            & 0.442          & 0.728          & 0.331          & 0.923          \\
        + Contrastive                             &            &            & \checkmark & 0.549          & 0.613          & 0.399          & \textbf{0.978} \\
        DANN + MMD                                & \checkmark & \checkmark &            & \textbf{0.637} & 0.665          & 0.412          & 0.891          \\
        All three                                 & \checkmark & \checkmark & \checkmark & 0.537          & \textbf{0.689} & \textbf{0.428} & 0.763          \\
        \bottomrule
    \end{tabular}
    \caption{Financial domain ablation (macro F$_1$). D\,=\,DANN, M\,=\,MMD, C\,=\,Contrastive. Each condition trains from scratch on financial as the sole target. $\dagger$: 8B values are means across 3 seeds, with high variance (std up to 0.118) due to the small 93-sample eval set.}
    \label{tab:ablation_financial}
\end{table*}

For SST-2, we ran the six-condition ablation across all four backbone families, as shown in Table~\ref{tab:ablation_sst2}. Every backbone exhibits the same flat pattern: all conditions fall within 0.006 of each other per model, the plain baseline is best or tied-best in every column, and the full combination is at or below baseline in all cases. This finding holds across a 13$\times$ range of model scale and across both general-purpose and domain-specialized backbones, confirming that DA provides negligible benefit on close-domain transfer regardless of capacity or objective.

\begin{table*}[t]
    \centering
    \normalsize
    \begin{tabular}{@{}lccccccc@{}}
        \toprule
        \multicolumn{1}{c}{\textbf{Condition}}    &
        \multicolumn{1}{c}{\textbf{D}}            &
        \multicolumn{1}{c}{\textbf{M}}            &
        \multicolumn{1}{c}{\textbf{C}}            &
        \multicolumn{1}{c}{\textbf{0.6B}}         &
        \multicolumn{1}{c}{\textbf{4B}}           &
        \multicolumn{1}{c}{\textbf{8B}$^\dagger$} &
        \multicolumn{1}{c}{\textbf{FinBERT}}                                                                                                                 \\
        \midrule
        Baseline                                  &            &            &            & 0.878          & \textbf{0.928} & 0.918          & \textbf{0.856} \\
        + DANN                                    & \checkmark &            &            & 0.877          & 0.910          & 0.919          & 0.848          \\
        + MMD                                     &            & \checkmark &            & \textbf{0.881} & 0.908          & \textbf{0.921} & 0.841          \\
        + Contrastive                             &            &            & \checkmark & \textbf{0.881} & 0.921          & 0.919          & 0.835          \\
        DANN + MMD                                & \checkmark & \checkmark &            & 0.876          & 0.915          & 0.919          & 0.850          \\
        All three                                 & \checkmark & \checkmark & \checkmark & 0.875          & 0.921          & 0.915          & 0.828          \\
        \bottomrule
    \end{tabular}
    \caption{SST-2 ablation (macro F$_1$). D\,=\,DANN, M\,=\,MMD, and C\,=\,Contrastive. Bold marks the best condition per backbone. The plain baseline leads or ties in every column, and DA shows no consistent benefit on close-domain transfer.}
    \label{tab:ablation_sst2}
\end{table*}

The financial ablation reveals three distinct patterns across backbone types.

\paragraph{Small general-purpose backbones (0.6B)}
Without DA, the adapter generalizes poorly to financial text (0.309), reflecting the backbone's limited financial pretraining. DA provides substantial recovery: DANN+MMD together yield the largest gain (+0.328 over baseline). However, adding contrastive loss to DANN+MMD reduces performance (0.537 vs.\ 0.637), suggesting a negative interaction when all three components operate simultaneously on a low-capacity backbone.

\paragraph{Larger general-purpose backbones (4B)}
The stronger backbone produces a better no-DA baseline (0.632). DANN and MMD each improve performance individually (+0.075 and +0.096, respectively), while contrastive loss alone slightly reduces performance (-0.019). DANN+MMD reaches 0.665, which is lower than MMD alone, indicating that the two alignment losses do not combine additively in this setting. The full combination is the best configuration for 4B (0.689), suggesting that contrastive regularization becomes beneficial at this capacity only when combined with the alignment losses.

\paragraph{Large general-purpose backbones (8B)}
At 8B scale, all six conditions cluster between 0.33 and 0.43, averaged across three seeds, with no condition clearly dominant. The high variance (std up to 0.118) and the small 93-sample financial evaluation set prevent drawing strong conclusions from these ablation values. Notably, a large gap exists between the ablation baseline (0.421) and the 8B main result (0.944). Unlike smaller models, the 8B backbone appears to benefit substantially from the curriculum training protocol, which adapts first to SST-2 and then to Financial PhraseBank. In contrast, the fresh-start financial ablation appears unable to fully exploit the backbone's representation quality. We therefore treat the 8B financial ablation as indicative rather than conclusive.

\paragraph{Domain-specialized backbones (FinBERT)}
FinBERT's no-DA baseline (0.902) is the highest cold-start result in our study, confirming that financial pretraining provides strong prior knowledge. The component interactions differ sharply from those observed for general-purpose backbones. DANN is actively harmful (-0.106): gradient reversal forces the adapter to make representations less distinguishable across financial and non-financial domains, erasing the domain-specific structure that makes FinBERT valuable. MMD provides a small benefit (+0.021). Contrastive loss produces the largest single-run improvement (+0.076) in our ablation, reaching 0.978 on the 93-example evaluation set, the best financial result we observe across configurations and models. This result has not been validated across multiple seeds and should be treated as preliminary. It sharpens class separation within the existing financial representation space without disrupting domain structure. The full combination (0.763) performs worse than no domain adaptation, driven by the destructive effect of DANN.

\begin{figure}[!htbp]
    \centering
    \begin{subfigure}[t]{\columnwidth}
        \includegraphics[width=\columnwidth]{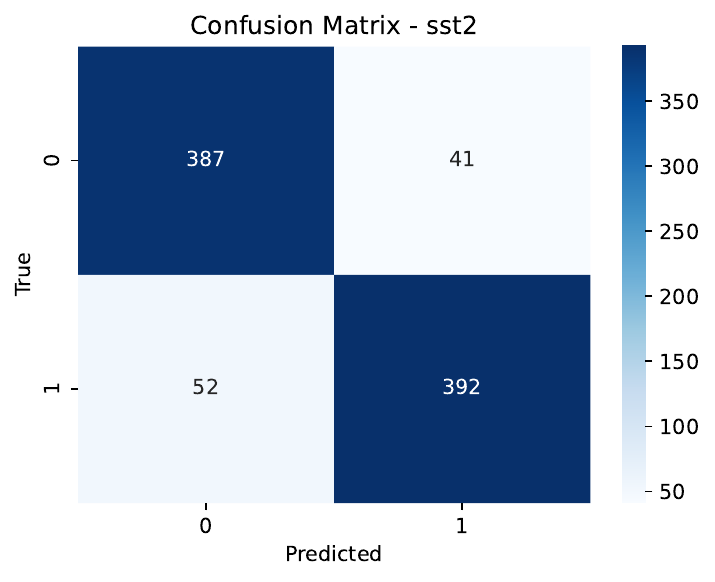}
        \caption{SST-2}
        \label{fig:cm_sst2}
    \end{subfigure}
    \begin{subfigure}[t]{\columnwidth}
        \includegraphics[width=\columnwidth]{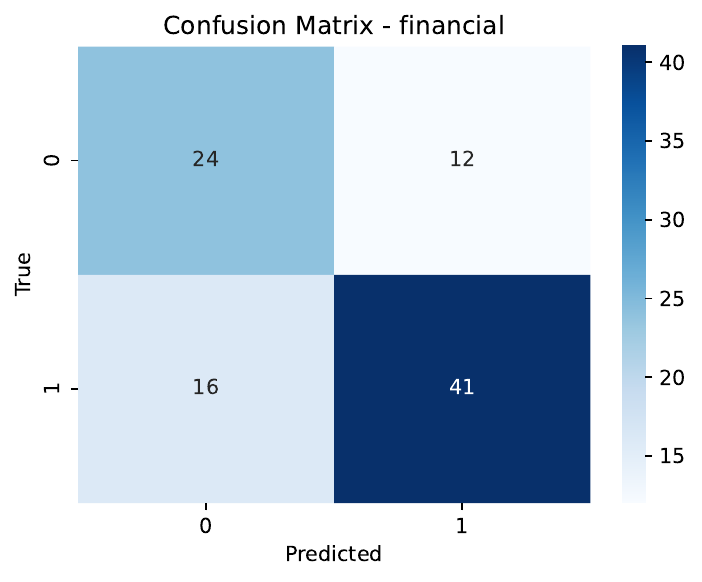}
        \caption{Financial}
        \label{fig:cm_financial}
    \end{subfigure}

    \caption{Confusion matrices for Qwen3-0.6B on zero-shot transfer targets.}
    \label{fig:cms}
\end{figure}

\subsection{Discussion}

Our results show that the benefit of domain adaptation is determined by one factor: whether the frozen backbone already covers the target distribution. On SST-2, consumer reviews and movie reviews share informal opinionated language and similar lexical sentiment cues, so all backbone configurations produce representations that are already linearly separable on the target. In this regime, DA objectives have little distribution gap to close, and the added regularization from DANN and MMD marginally degrades class boundaries rather than improving them, explaining the flat ablation pattern across all conditions.

Financial PhraseBank presents a fundamentally different setting. Its formal, entity-centric language relies on domain-specific vocabulary that is largely absent from consumer review corpora, placing source and target representations in different regions of feature space. For small general-purpose backbones, explicit distribution matching with DANN and MMD appears to bridge this gap, recovering up to 32.8 macro F$_1$ points for the 0.6B backbone in our single-run evaluation. This is consistent with the regime these objectives were designed for.

The FinBERT results reveal a fundamental incompatibility between adversarial alignment and frozen domain-specialized backbones. DANN is designed to encourage a jointly trained feature extractor to discard domain-specific structure. When the backbone is frozen, however, that structure is fixed and cannot be re-learned within the backbone. Gradient reversal therefore forces the adapter to suppress informative financial features, weakening the specialized representation manifold and reducing macro F$_1$ by 0.106. Supervised contrastive loss avoids this failure mode. Instead of aligning source and target distributions, it sharpens sentiment boundaries within the existing representation space while preserving financial-domain structure. This yields 0.978 macro F$_1$, the best result we observe across all configurations. These observations suggest a candidate selection heuristic worth further validation: use distribution-matching DA when the backbone lacks target-domain coverage, and use contrastive refinement when target-domain structure is already present.

This exploratory setup is subject to several important limitations. First, our focus on binary sentiment and product/restaurant source domains means these trends may not generalize to multi-class tasks or other text types. Second, domain adaptation loss weights are held fixed across all configurations, which precludes model-specific optimization. Third, our Financial PhraseBank evaluation set is relatively small, which induces high variance in some configurations. For example, the 8B model has a standard deviation of 0.118 across seeds. As a result, these results should be interpreted as directional trends rather than definitive benchmarks. Finally, the main pipeline's sequential curriculum, which adapts to SST-2 before financial text, confounds direct comparison with our fresh-start ablations. The substantial performance divergence between these two settings for large-capacity models suggests that sequential target exposure significantly alters adapter convergence, a dynamic that warrants isolated study in future work.

\section{Conclusion}
\label{sec:conclusion}

This work examines when domain adaptation improves sentiment transfer with frozen pre-trained backbones. By keeping the backbone fixed and training only a lightweight adapter, we isolate the effect of adaptation objectives across backbone scale and domain specialization.

Our observations point to two transfer regimes. On SST-2, the close distributional proximity to the source domains makes explicit adaptation unnecessary across all backbone scales in our setting. On Financial PhraseBank, distribution-matching objectives recover substantial performance for general-purpose backbones, while supervised contrastive learning is the safest and most effective objective for domain-specialized backbones such as FinBERT in our experiments. Taken together, these case-study observations support a backbone-aware DA selection strategy that matches the adaptation objective to the representation gap rather than applying a fixed recipe. However, confirming this as a general strategy will require validation on larger, multi-seed benchmarks and additional source-target domain pairs.

\bibliographystyle{unsrt}
\bibliography{ref}

\end{document}